\def\year{2020}\relax
\documentclass[letterpaper]{article} % DO NOT CHANGE THIS
\usepackage{aaai20}  % DO NOT CHANGE THIS
\usepackage{times}  % DO NOT CHANGE THIS
\usepackage{helvet} % DO NOT CHANGE THIS
\usepackage{courier}  % DO NOT CHANGE THIS
\usepackage[hyphens]{url}  % DO NOT CHANGE THIS
\usepackage{graphicx} % DO NOT CHANGE THIS
\usepackage{graphicx}  % DO NOT CHANGE THIS
\title{Challenges and Opportunities for Computer Vision in Real-life Soccer Analytics}

\author{Neha Bhargava, Fabio Cuzzolin\\
Oxford Brookes University, Oxford, UK\\
nbhargava@brookes.ac.uk, fabio.cuzzolin@brookes.ac.uk
}

\begin{document}
\maketitle

\begin{abstract}
In this paper, we explore some of the applications of computer vision to sports analytics. Sport analytics deals with understanding and discovering patterns from a corpus of sports data. Analysing such data provides important performance metrics for the players, for instance in soccer matches, that could be useful for estimating their fitness and strengths. Team level statistics can also be estimated from such analysis. This paper mainly focuses on some the challenges and opportunities presented by sport video analysis in computer vision. Specifically, we use our multi-camera setup as a framework to discuss some of the real-life challenges for machine learning algorithms.

\end{abstract}

\section{Introduction}
%The industry of sport analytics (especially in \textit{football}) is growing exponentially owing to its increasing market. 
Research in sports analytics has recently substantially increased because of the availability of a huge corpus of data. Such data provides a challenging test-bed for machine learning algorithms - e.g. for tracking, action and activity recognition etc. At the same time, the huge commercial interests in a better understanding of player's and team's abilities using sport analytics is encouraging a lot of interest in the field. Team sport analytics deals with the analysis of long-term data of both individual players and teams. The most common data forms are GPS tracks and videos. Such an analysis can assist the clubs, coaches, and players in decision making - at player level and at team level. At player level, the individual statistics can assist in assessing one's performance, fitness level, strengths and weaknesses, etc. At team level, it could assist in team building, tactical analysis, formation planning, etc. This paper focuses on soccer in particular, and discusses the challenges and opportunities available for the fields of computer vision and machine learning in this sport. \\

Tracking players during matches and training sessions is of high importance because numerous performance metrics (e.g. high-speed runs, acceleration, deceleration, etc.) can be extracted from these tracks. These metrics are useful to sports scientists in accessing a player's fitness, strengths and other factors. There are different ways of tracking players - using wearable sensors such as GPS or using camera(s). Once the tracks are available, these  different metrics can be estimated. These metrics are based solely on track data, hence they may not be enough to provide the complete profile of a player. For example, \textit{jumping} is an important ability for attacking players as well as for defending players, whereas tracking data cannot really quantify such features because of its inability to identify such events or actions. The ability to capture different actions performed by a player either in a match or during a training session can enhance the understanding of the overall performance and importance of the each player in the team. Additionally, to understand the game at a higher level, one needs to know what each player is doing at any given point of time and understand player interactions over the time. This is where computer vision can pitch in to contribute. Developing algorithms for recognising a single player's actions, multiple players interactions, and team tactics can be a step towards a complete understanding of the match. \\

To capture visual data of the players during the match or a training session, each club or sport analytics company has its own unique setup. Some would use multiple cameras around the  field while others would use a single panoramic camera. These different setups pose different challenges and need different approaches to address them. The primary objective of this paper is to describe and discuss some of the challenges that are common in a real-life sport analytics setup. Towards this, we discuss the unique setting we are working with. Unfortunately, due to privacy regulation, we cannot release any image in the paper from our dataset. We organise the paper in the following way. After a brief note on related work, we discuss the input format. We then formally present our problem statement. We list out the challenges associated with the setup followed by a few specific challenges in sport action detection. We provide some experimental results and discussions followed by our conclusions. 

\section{Related Work}
Sport analytics has recently gained massive attention from the AI researchers.
%, primarily because of two reasons - (a) it provides a challenging and real-life test-bed for machine learning algorithms, (b) existence of a potentially huge commercial market. 
One of the most frequently used techniques in sports analysis is tracking. Player tracking is useful in estimating performance metrics for the player \cite{pt2,pt3}. Ball tracking is important to analyse ball possession statistics \cite{bt3,bt4}. Also, there has been some work in automatic understanding of sports videos \cite{ana3,ana4}. Nevertheless, there is a lot of scope in hierarchical understanding of a match.\\

Sport action detection is a problem of classifying an action as well as localising it temporally and spatially in the input video. It is a widely studied problem in computer vision. Action detection models can be either single frame based \cite{det_frame_2,det_frame_3} or multi-frame based \cite{tube3,tube4}. Action recognition is a relatively simpler task of predicting a class label for an input video. Some of the single frame based action recognition models are proposed in \cite{rec_frame_1,rec_frame_2}. There is also a substantial amount of work on video based models \cite{rec_vid_1,rec_vid_3}.  \\

Additionally, several sports datasets are available to be used as benchmarks for sports analysis, namely \textit{UCF101}\cite{ucf101}, \textit{Sports Videos in the Wild (SVW)}\cite{svw}, \textit{Sports-1M}\cite{rec_vid_1}, \textit{SoccerNet}\cite{soccernet}. Out of all these datasets, \textit{UCF101}, \textit{SVW}, and \textit{Sports-1M} are generic in the sense that they contain videos from multiple sports. \textit{SoccerNet} is specific to soccer but contains annotations for limited events. The lack of sport specific datasets with extensive annotations poses some limitations in learning sport specific actions.

\section{Input Setup}
In this section, we describe our input setup. We have a multiple cameras framework. Each camera has a frame-rate of 10 fps and provides thumbnails of players that are in its field of view. A thumbnail is a cropped part in the image containing a player and the surrounding.  %the player captures ${1/6}^{th}$ of the thumbnail.
That is, if a camera sees $M$ players at time $t$, we extract $M$ thumbnails from the camera image for the time instant $t$. The remaining part of the image is discarded and not saved due to memory constraints. %These thumbnails have very low resolution. To add to the complexity, cameras may sometime capture frames at a rate that varies with time, or discard some video frames entirely. So in essence, we have thumbnails of the players at irregular times and from different cameras. 
The following points summarise the setup:  
\begin{itemize}
    \item If a player is visible in a camera for a duration, the camera produces thumbnails around the player for the duration. Since the camera may drop frames in between, these thumbnails are produced at irregular time intervals.
    \item If a player is visible in more than one camera, we have multiple thumbnails for the player from different cameras. 
    \item The resolution of the thumbnails is $256\times256$. 
\end{itemize}

%In summary, we have multiple low-resolution thumbnails of the players that are coming from varying number of cameras with irregular time stamps.

\section{Problem Statement}
\begin{figure}
    \centering
    \includegraphics[scale=0.25]{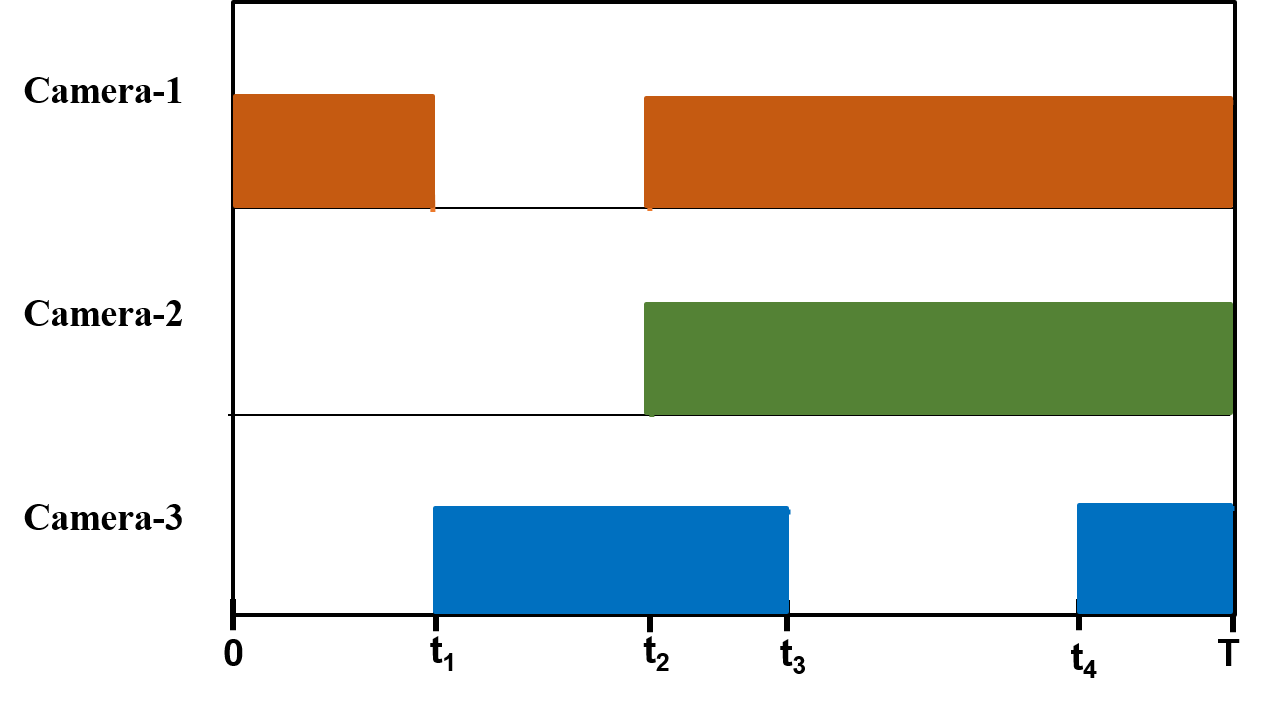}
    \caption{Illustration of data capture process under multi-camera setup. The coloured boxes denote the appearance of a player in the corresponding camera for that duration.}
    \label{fig:setup}
\end{figure}

Just to recall the input setup in a formal way, let there be $N$ cameras and $M$ players. The video is recorded over a time interval $[0,T]$. Consider the $i^{th}$ player for this duration. This player is visible $K_j^i$ times in $j^{th}$ camera. Hence we have $K_j^i$ sequences of thumbnails for $i^{th}$ player that come from $j^{th}$ camera with time stamps of $[t^{i}_{1j}, t^{i}_{2j}]^k$ where $k~\forall~[1, K_j^i]$. Figure~\ref{fig:setup} illustrates this setup with a toy example. Consider that we have three cameras in our setup. These cameras capture thumbnails of a player at different times. The coloured boxes denote the appearance of a player in the corresponding camera. For example, (a) Camera one sees the player from periods $[0,t_1]$ and $[t_2,T]$, (b) the player is visible in all the cameras in the periods $[t_2,t_3]$ and $[t_4,T]$. When more than one player is present on the field, then for each camera and for each player all the thumbnails need to be associated in time to create the relevant tracklets.
%When there are more than one player, we need to associate these thumbnails for a camera to get such tracklets for each player.\\

We are interested in solving the following problems under this challenging and unique data collection setup - 
\begin{enumerate}
    \item Thumbnails can contain other players also. So it is essential to localise the players in the thumbnails and identify the central player whom this thumbnail belongs to. 
    \item A camera generates thumbnails of the players. We need to associate these thumbnails to generate the tracklets of the players as seen by the camera. 
    \item Combine the tracklets of the players from all the cameras to track them over the whole time interval $[0,T]$. %In an extreme situation when the players are visible in all the cameras for the complete duration, ideally we would get $M$ tracks for $N$ cameras.
    \item Number detection: The number printed on a player's jersey is useful to associate two tracklets. Hence, it is an important problem.
    \item Recognise the sequence of actions performed by each player over this time duration. A few examples of individual actions that potentially of interest in the sports analytics domain, are \textit{jumping, kicking, running, etc.}
    \item Recognise the activities that are performed by multiple players in collaboration. A few examples of interactions are \textit{passing a ball, tackling, etc.}
\end{enumerate}

Out of these interesting problems, we discuss only a couple of problems in the paper as a part of our initial experimentation $-$ player detection and number detection. In the next section, we discuss some of the challenges that need to be addressed in order to be able to solve these problems.

\begin{figure}
    \centering
    \includegraphics[scale=1.55]{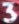}~
    \includegraphics[scale=1.45]{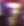}~
    \includegraphics[scale=1.1]{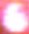}~
    \includegraphics[scale=1.4]{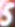}~
    \includegraphics[scale=1.2]{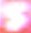}~
    \includegraphics[scale=1.2]{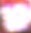}
    \caption{Illustration of wide range of image quality in our dataset.}
    \label{fig:imgs}
\end{figure}

\section{Challenges}
The unique setup poses some unique challenges and questions as well. In this section, we discuss some of them:
\begin{enumerate}
    \item The base frame-rate of the cameras is low. How difficult is it to capture the fast actions with slow frame-rate?
    \item The existing models usually work with fixed frame-rate videos. In our case, sometimes the frames are dropped intermittently. So an interesting challenge is to address the varying nature of the frame-rate with such models. 
    \item The quality of the thumbnails are poor, making the computer vision tasks harder to solve effectively. Figure \ref{fig:imgs} shows a few examples from our dataset. The images are cropped because of privacy issues.
    \item Since the players are highly mobile in the game, it is possible that the parts of the same action are visible in different cameras. The challenge is to combine the relevant clips from different cameras to identify the action performed.
    \item Is it possible to identify the interactions at all in such a setup where the field context is not available?
    %\item We have limited labelled data for all these tasks. How to address all these challenges with small data?
\end{enumerate}
Next, we look at some of the challenges pertaining to sport action detection task.

\begin{table}[]
\resizebox{\columnwidth}{!}{%
\begin{tabular}{|l|l|l|l|l|l|l|l|l|l|l|}
\hline
\textbf{Classes} & 0    & 1    & 2    & 3    & 4    & 5    & 6    & 7    & 8    & 9    \\ \hline
\textbf{mAP}     & 0.57 & 0.57 & 0.68 & 0.61 & 0.30 & 0.58 & 0.47 & 0.31 & 0.29 & 0.51 \\ \hline
\end{tabular}%
}
\caption{Class-wise mAP on validation dataset for digit detection.}
\label{table:mAP_digits}
\end{table}

\begin{table}[]
\resizebox{\columnwidth}{!}{%
\begin{tabular}{|l|c|c|c|c|}
\hline
\multicolumn{1}{|c|}{\textbf{Model}}                                                    & \textbf{Time}  & \textbf{AP\_{[}0.50:0.95{]}} & \textbf{AP\_0.50} & \textbf{AP\_0.75} \\ \hline
Faster-RCNN-Inception                                                                   & 51 ms          & 0.45                         & \textbf{0.72}     & 0.50              \\ \hline
\begin{tabular}[c]{@{}l@{}}Faster-RCNN-Inception-Resnet \\ (300 proposals)\end{tabular} & 345 ms         & 0.43                         & 0.68              & 0.49              \\ \hline
\begin{tabular}[c]{@{}l@{}}Faster-RCNN-Inception-Resnet\\ (20 proposals)\end{tabular}   & 109 ms         & 0.36                         & 0.54              & 0.41              \\ \hline
RetinaNet                                                                               & \textbf{35 ms} & 0.20                         & 0.45              & 0.14              \\ \hline
\end{tabular}%
}
\caption{Performance comparison of some object detectors on the test-set. RetinaNet was the fastest but the performance was poor. Faster-RCNN-Inception was found to be optimal in term of speed and performance.}
\label{table:mAP_person}
\end{table}

\section{Sport Action Detection}
In sports analytics, automatic detection or recognition of a sequence of single player actions and multi-player interactions can provide useful insights. Since we do not have a typical input setting, we need to investigate the following:
\begin{enumerate}
    \item Typically, the existing methods take fixed frame-rate videos as an input. It would be interesting to test their applicability on variable frame-rate videos.
    \item The base frame-rate is 10 fps and at times, it can go further down. Some investigation is required to see if the existing methods can capture the fast actions with this frame-rate as well.
    \item Recall that we have thumbnails of individual players from the different cameras. It is challenging to recognise the multi-player interactions in such a setup. 
\end{enumerate}
Our work programme is to address these challenges. We started with the tasks of player detection and jersey number detection. In the next section, we discuss the experimental details.

\section{Evaluation}
\begin{figure}
    \centering
    \includegraphics[scale=0.5]{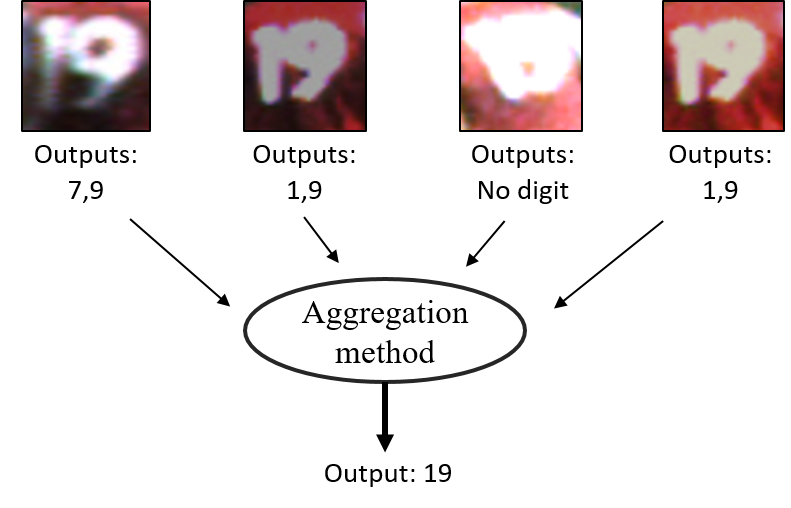}
    \caption{Aggregation of multiple predictions that are coming from different thumbnails to get the final jersey number for the player.}
    \label{fig:agg}
\end{figure}

As mentioned, we started tackling the relatively simpler tasks of number and player detection. For these tasks, we tested a few object detectors.
\begin{enumerate}
    \item \textbf{Number detection}: The objective is to identify the number printed on player's jersey from a set of thumbnails of the player. We addressed the number detection problem into two stages - (a) Digit detection, and (b) Aggregation of all the predicted digits. To detect the digits, we used {RetinaNet} \cite{retinanet} with data augmentation. We pre-train the model using \textit{SVHN} dataset \cite{svhn} which consists of house numbers. The mAP on \textit{SVHN} validation set was 0.92. We subsequently fine-tune the model with our dataset. The training and validation datasets consist of around 10,500 images and 2,800 images, respectively. The training dataset was unbalanced, so we used class weights to lessen its effects. We achieved an mAP of 0.48 on our validation set. The class-wise mAP is mentioned in Table \ref{table:mAP_digits}. We suspect the reason for the performance gap on our and \textit{SVHN} datasets is the poor quality of our images, as shown in Figure \ref{fig:imgs}. \\
    
    The next task is to combine all the predictions and get a final number for the player. Some of the challenges in the task are - missing predictions (e.g. no prediction in a thumbnail), partial predictions (e.g. a digit is missing from a number), presence of multiple numbers in the thumbnails, etc. We use Dempster's rule for combination of multiple evidences (which is based on Dempster-Shafer Theory) \cite{dst}, where each thumbnail provides some evidence for the jersey number. An illustration of an example is shown in Figure \ref{fig:agg}. The different outputs from all the thumbnails are aggregated to estimate the final number for the player.
     
    \item \textbf{Player detection}: The objective is to localise the players in an input thumbnail. This is useful in player matching for the task of tracklet generation for the players, and for action recognition. We tested recent object detectors - Faster-RCNN \cite{faster_rcnn} (Inception based), Faster-RCNN (ResNet-Inception based), and RetinaNet \cite{retinanet}. We fine-tune these models with our dataset but did not do any parameter tuning. The raw performance of these detectors on our test-set is mentioned in the Table \ref{table:mAP_person}. Faster-RCNN (Inception) performed the best with respect to mAP. RetinaNet was the fastest one but the performance was poor. We fine-tune the parameters of Faster-RCNN (Inception) to achieved a bit higher mAP of 0.74.  
    
\end{enumerate}

%These initial and small set of experiments inspired us to reflect on our standard approach to fine tuning and other aspects. We discuss a few of them in the next section.

\section{Discussions}
These initial and small set of experiments inspired us to reflect on our standard approach to fine tuning and other aspects. Many of such related concerns are already under investigation by many researchers and have given rise to interesting directions.
\begin{enumerate}
    \item Do state-of-the-art methods perform as well on real life data as they perform on benchmark datasets? The performance of {RetinaNet} for player detection task was disappointing and hence raised a few concerns about the benchmark datasets - Are they skewed? The real-life dataset comes with real-life variety. For example in soccer, the videos could have - noise due to weather conditions; the motion blur due to fast movement of players; variable size of the player and the ball depending on their distances from the cameras; considerable amount of occlusion due to multi-player interactions. Do the benchmark datasets have enough variability to judge the generalised performance of a model? How do we quantify the amount of real-life variety in benchmark datasets? Maybe we need to examine our benchmark datasets in more principled way.
    \item In both of our experiments, we initially fine-tuned the classifier layer only and not the feature extractor but that led to poor performance of the model. We achieved better performance after fine-tuning the feature extractor. So one natural question arises about what is an optimal and principled transfer learning approach for real-life datasets.   
\end{enumerate}

\section{Conclusions}
In this paper, we discussed the importance of data analytics in soccer and the role of AI. We detailed some of the challenges that we faced in our initial experiments. Our experimentation raised a few concerns regarding the benchmark datasets and the applicability of state-of-the-art methods on real-life problems. Finally, we explored the opportunities in the form of questions that are lying ahead of us in the field of sports analytics which truly provides a challenging and real-life test-bed.

\end{document}